%% file: MAIN.tex
\documentclass[shortAfour,sageh,times]{sagej}
\usepackage{moreverb,url}
\usepackage[colorlinks,bookmarks=false,citecolor=blue,urlcolor=blue]{hyperref}

\usepackage{epstopdf}
\usepackage{amsmath}
\usepackage{amssymb}
\usepackage{latexsym}
\usepackage{url}
\usepackage{relsize}
\usepackage{multirow}
\usepackage{afterpage}
\usepackage{ifthen}
\usepackage{graphicx}
\usepackage{algpseudocode,algorithm,algorithmicx}
\usepackage{subfigure}
\usepackage{flushend}
\usepackage{epstopdf}
\usepackage{stackengine}
\usepackage{gensymb}
\usepackage{booktabs}
\usepackage{makecell}
\usepackage{hhline}
\usepackage{courier}
\usepackage{natbib}
\usepackage{titlesec}
\usepackage{lipsum} 
\usepackage[figurename=Fig.]{caption}  

\usepackage{makecell}
\usepackage[10pt]{moresize}
\usepackage{anyfontsize}

\hypersetup{pdfstartview={FitH}}
\setcitestyle{authoryear,open={(},close={)}}
\setcitestyle{aysep={,}} 
\newcommand{\ra}[1]{\renewcommand{\arraystretch}{#1}}

\titleformat*{\section}{\large\bfseries\fontsize{12}{12}\selectfont} 
\titleformat*{\subsection}{\normalsize\itshape\fontsize{11}{11}\selectfont}
\titleformat*{\subsubsection}{\normalsize\itshape}

\makeatletter 
\renewcommand{\@seccntformat}[1]{\csname the#1\endcsname\ }  
\makeatother

\newcommand \tableCaption[6]{ 
    \begin{tabular}{l@{\hspace{2.0\tabcolsep}}l} 
    \scriptsize  \textbf{GT}: #1 		&	\scriptsize \textbf{SGC}: #3\\     
    \scriptsize  \textbf{EDNet}: #2 	&	\scriptsize \textbf{S2VT}: #4\\    
    \scriptsize  \textbf{V2CNet}: #5 	&	\scriptsize \textbf{SCN}: #6\\    
     \end{tabular}
}

\newcommand \datasetCaption[2]{ 
    \begin{tabular}{ll} 
         	\scriptsize \textbf{Command}: #1 	&	\scriptsize \textbf{Action}: #2     
         	\vspace{1.5ex}
     \end{tabular}
}

\usepackage{color, colortbl}
\definecolor{Gray}{gray}{0.9}

\newcommand\BibTeX{{\rmfamily B\kern-.05em \textsc{i\kern-.025em b}\kern-.08em
T\kern-.1667em\lower.7ex\hbox{E}\kern-.125emX}}
\def\volumeyear{2018}
\begin{document}
\input{0_localmacros}

\runninghead{Nguyen et al.}

\title{\Large V2CNet: A Deep Learning Framework to Translate Videos to Commands for \\ Robotic Manipulation}

\author{\normalsize  Anh Nguyen\affilnum{1}, Thanh-Toan Do\affilnum{2}, Ian Reid\affilnum{2}, Darwin G. Caldwell\affilnum{1}, Nikos G. Tsagarakis\affilnum{1}}

\affiliation{\affilnum{1} Department of Advanced Robotics, Istituto Italiano di Tecnologia, Via Morego 30, 16163,
Genova, Italy. \\ \affilnum{2} Australian Centre for Robotic Vision, The University of Adelaide, SA 5005, Australia.}
\corrauth{Anh Nguyen, Istituto Italiano di Tecnologia (IIT), Via Morego 30, 16163, Genova, Italy.}
\email{anh.nguyen@iit.it}


\begin{abstract}
\normalfont
\itshape
We propose V2CNet, a new deep learning framework to automatically translate the demonstration videos to commands that can be directly used in robotic applications. Our V2CNet has two branches and aims at understanding the demonstration video in a fine-grained manner. The first branch has the encoder-decoder architecture to encode the visual features and sequentially generate the output words as a command, while the second branch uses a Temporal Convolutional Network (TCN) to learn the fine-grained actions. By jointly training both branches, the network is able to model the sequential information of the command, while effectively encodes the fine-grained actions. The experimental results on our new large-scale dataset show that V2CNet outperforms recent state-of-the-art methods by a substantial margin, while its output can be applied in real robotic applications. The source code and trained models will be made available.

\end{abstract}


\maketitle

\input{1_intro}

\input{2_rw}

\input{3_rnn}

\input{4_exp}

\input{5_discussion}

\input{6_conclusions}

\bibliography{reference}

\end{document}

%% file: 0_localmacros.tex
\newcommand{\junk}[1]{}

%% file: 1_intro.tex
\section{Introduction} \label{Sec:Intro}
While humans can effortlessly understand the actions and imitate the tasks by just \textit{watching} someone else, making the robots to be able to perform actions based on observations of human activities is still a major challenge in robotics~\citep{Chrystopher09}. By understanding human actions, robots may acquire new skills, or perform different tasks, without the need for tedious programming. It is expected that the robots with these abilities will play an increasingly more important role in our society in areas such as assisting or replacing humans in disaster scenarios, taking care of the elderly, or helping people with everyday life tasks. Recently, this problem has been of great interest to researchers, many approaches have been proposed to tackle different tasks such as pouring water~\citep{pastor2009learning}, drawer opening~\citep{rana2017towards}, and multi-stage manipulation~\citep{zhang18_vr_manipulation}.

In this work, we argue that there are two main capabilities that a robot must develop to be able to replicate human activities: \textit{understanding} human actions, and \textit{imitating} them. The imitation step has been widely investigated in robotics within the framework of learning from demonstration (LfD)~\citep{Brenna2009}. In particular, there are two main approaches in LfD that focus on improving the accuracy of the imitation process: kinesthetic teaching~\citep{Akgun2012} and motion capture~\citep{Koenemann2014}. While the first approach needs the users to physically move the robot through the desired trajectories, the second approach uses a bodysuit or camera system to capture human motions. Although both approaches successfully allow a robot to imitate a human, the number of actions that the robot can learn is quite limited due to the need of using expensively physical systems (i.e., real robot, bodysuit, etc.) to capture the training data~\citep{Akgun2012, Koenemann2014}. 

The \textit{understanding} step, on the other hand, receives more attention from the computer vision community. Two popular problems that receive a great deal of interest are video classification~\citep{Karpathy2014} and action recognition~\citep{Simonyan2014}. Recently, with the rise of deep learning, the video captioning task~\citep{Venugopalan2016} has become more feasible to tackle. Unlike the video classification or detection tasks which output only the class identity, the output of the video captioning task is a natural language sentence, which is potentially useful in robotic applications.

In this paper, we propose a new deep learning framework that translates a demonstration video to a command which can directly be used in robotic applications. While the field of LfD focuses mainly on the imitation step, we focus on the understanding step, but our proposed method also allows the robot to perform useful tasks via the output commands. Our goal is to bridge the gap between computer vision and robotics, by developing a system that helps the robot understand human actions, and use this knowledge to complete useful tasks. Furthermore, since our method provides a meaningful way to let the robots understand human demonstrations by encoding the knowledge in the video, it can be integrated with any LfD techniques to improve the manipulation capabilities of the robot.

In particular, we cast the problem of translating videos to commands as a video captioning task: given a video, the goal is to translate this video to a command. Although we are inspired by the video captioning field, there are two key differences between our approach and the traditional video captioning task: (i) we use the \textit{grammar-free} format in the captions for the convenience in robotic applications, and (ii) we aim at learning the commands through the demonstration videos that contain the \textit{fine-grained} human actions. The use of fine-grained classes forms a more challenging problem than the traditional video captioning task since the fine-grained actions usually happen within a short duration, and there is usually ambiguous information between these actions. To effectively learn the fine-grained actions, unlike the traditional video and image captioning methods~\citep{yao2015describing, you16_image_captioning} that mainly investigate the use of \textit{visual attention} to improve the result, we propose to use the Temporal Convolutional Networks (TCN) to explicitly classify the human actions. This TCN network can be considered as an \textit{action attention} mechanism since it does not focus on the visual part of the data, but deeply encodes the fine-grained actions that humans are performing in the videos.

\begin{figure}[!t] 
    \centering
     \includegraphics[scale=0.265]{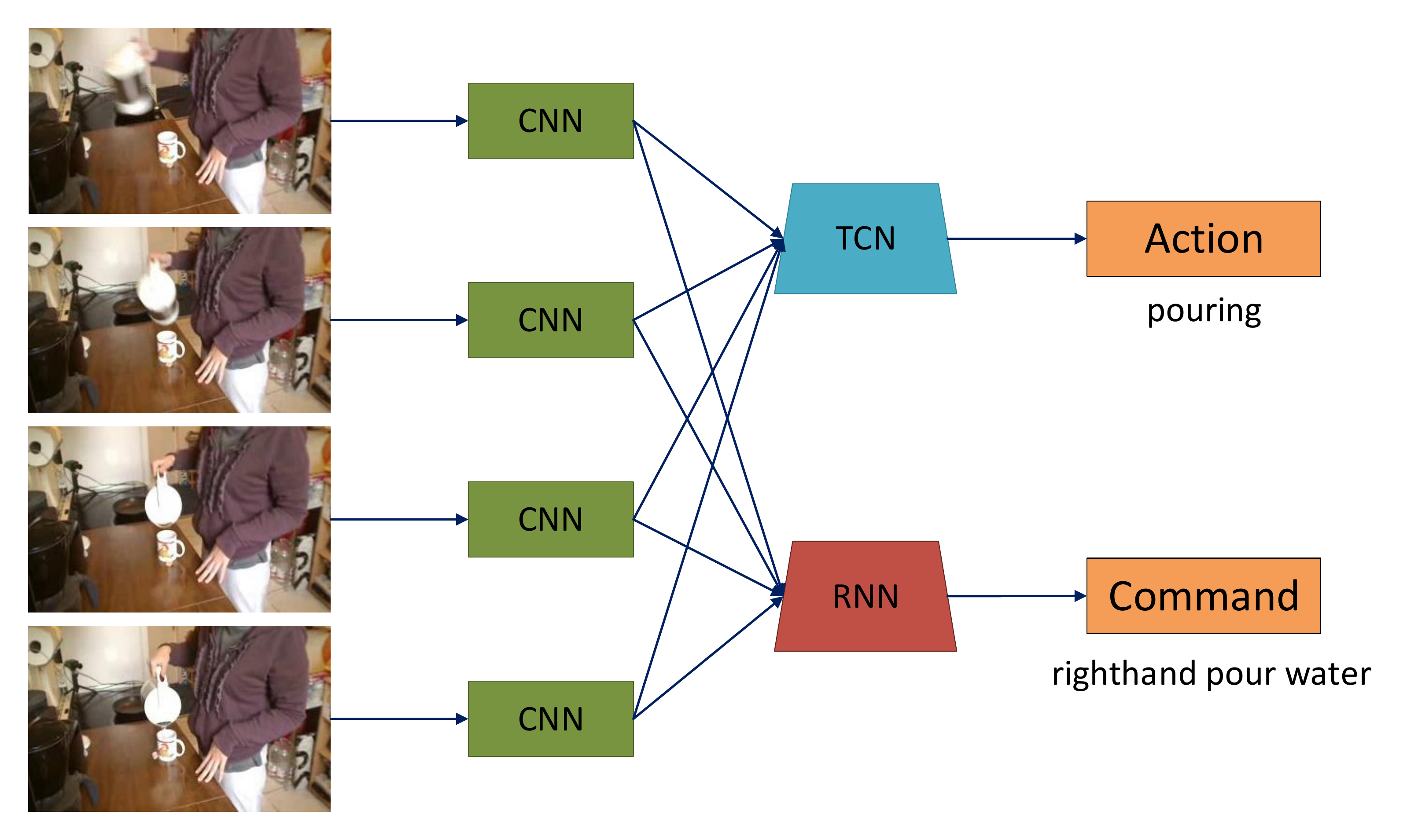}   
    \vspace{-3.0ex}
    \caption{An overview of our V2CNet architecture. The network is composed of two branches: a TCN branch to encode the actions, and a RNN branch to translate videos to commands.}
    \label{Fig:intro} 
\end{figure}

Based on the aforementioned observation, we design a network with two branches: a translation branch that interprets videos to commands using a RNN network, and an action classification branch that classifies the fine-grained human actions with a TCN network. The intuition of our design is that since both the classification and translation branches are jointly trained using a single loss function, the parameters of both branches are updated using the same gradient signal through backpropagation. Therefore, the classification branch will encourage the translation branch to generate the correct fine-grained action, which is the key information to understand human demonstrations. We experimentally demonstrate that by jointly train both branches, the translation results are significantly improved. Fig.~\ref{Fig:intro} illustrates an overview of our video-to-command (V2CNet) deep network.

To summarize, our main contributions are as follows:
\begin{itemize}
\item We reformulate the understanding step of the robot imitation problem as a fine-grained video captioning task, which potentially allows the robot to directly do the tasks from demonstration videos.
\item We present a new fairly simple, yet effective deep architecture that translates videos to commands.
\item We introduce a new large-scale dataset for the video-to-command task, which is suitable for deep learning approach.
\item The methodology of our approach is validated in the real robotic systems. The source code and trained models will also be released. 
\end{itemize}

Next, we review the related work in Section~\ref{Sec:rw}, then the detailed network architecture is described in Section~\ref{Sec:rnn}. In Section~\ref{Sec:exp}, we present the extensive experiments on our new IIT-V2C dataset and on the real robotic platforms. We discuss the advantages and limitations of our approach in Section~\ref{Sec:discussion}, then conclude the paper in Section~\ref{Sec:con}.

\junk{
Furthermore, since our method encodes the information of the demonstration video into a command,

, the network is able to effectively encode both the visual features and the fine-grained action, hence

While the video captioning problem describes the output sentence in a natural language form, we use a grammar-free form to describe the output command, which can be directly used in  real-world robotic applications.

r. The design of LSTM-TSA is highly inspired by the facts
that 1) semantic attributes play a significant contribution to
captioning, and 2) images and videos carry complementary
semantics and thus can reinforce each other for captioning

. We demonstrate that our solely neural architecture further improves the state of the art, while its output can be applied in real robotic applications.

   bran our goal is to generate an output command given an input video we first use CNN to extract deep features from video frames, then two RNN layers are used to learn the relationship between the visual features and the output command. 

The ability to perform actions based on observations of human activities is one of the major challenges to increase the capabilities of robotic systems~\citep{Chrystopher09}.

However, the outputs of these problems are discrete (e.g., the action classes used in~\citep{Simonyan2014} are ``diving", ``biking", ``skiing", etc.), and do not provide further meaningful clues that can be used in robotic applications.

Inspired by the recent advances in computer vision,

 actions from studying

The remaining of this paper is organized as follows.
that can be used directly in robotic experiments

directly

the state-of-the-art

discuss the advantages and limitations of our approach then

\begin{itemize}
\item A deep learning based method that can be trained end-to-end to directly interpret input videos to robot commands.

\item A dataset that is sufficient for deep learning methods.

\item An robotic framework that use the proposed method to execute different manipulation tasks.

\end{itemize}

This paper describes the implementation of the ChainModel [15], [16], a biologically inspired model of the mirror neuron system, on an anthropomorphic robot. The work focuses on an interaction and imitation task between the robot and a human, wheremotor control applied to object manipulation is integrated with visual understanding capabilities.More specifically, the re- search focuses on a fully instantiated system integrating perception and learning, capable of executing goal directed motor se- quences and understanding communicative gestures. In particular, the goal is to develop a robotic system that learns to use its motor repertoire to interpret and reproduce observed actions under the guidance of a human user.

A popular approach to solve this problem is learning from demonstration.

The framework of LfD is divided into two fundamental phases: motion transfer and skill modeling. There are two widely used approaches to transfer a motion: kinesthetic guiding and motion capture [4]. The former physically moving the robot in question. This is an intuitive way for users to teach motions to a robot and there are no correspondence issues since the robot is typically aware of its current configuration, i.e., through joint encoders [4,5]. Resulting behaviors tend to be unnatural, because kinesthetic guiding cannot exactly imitate a skill the same way humans would naturally perform it, and it may be difficult for the human to smoothly manipulate the robot. Alternatively, in motion capture systems, users demonstrate a behavior which is extracted by the system, and then converted to a suitable form after some pre-processing steps to remove noise in order to project it onto the robot. One of the main challenges of this approach is how to project recorded data onto the body of the robot, since its morphology may be different from that of the human – the correspondence problem [4]. Typically a body- suit system [6,7] or a camera-based system [8,9] is employed as a motion capturing system. The latter is preferable for demonstrators because it does not constrain performances, allowing for more natural or complex demonstrations.

Transferring skills to humanoid robots based on observations of human activities is widely considered to be one of the most effective ways of increasing the capabilities of such systems [1,2]. It is expected that semantic representations of human activities will play a key role in advancing these sophisticated systems beyond their current capabilities, which will enable these robots to obtain and determine high level understanding of human behaviors. The ability to automatically recognize a human behavior and react to it by generating the next probable motion or action according to human expectations will substantially enrich humanoid robots.

The ability to learn actions from human demonstrations is one of the major challenges for the development of intelligent systems. Particularly, manipulation actions are very challenging, as there is large variation in the way they can be performed and there are many occlusions

This paper describes the implementation of the ChainModel [15], [16], a biologically inspired model of the mirror neuron system, on an anthropomorphic robot. The work focuses on an interaction and imitation task between the robot and a human, wheremotor control applied to object manipulation is integrated with visual understanding capabilities.More specifically, the re- search focuses on a fully instantiated system integrating perception and learning, capable of executing goal directed motor se- quences and understanding communicative gestures. In particular, the goal is to develop a robotic system that learns to use its motor repertoire to interpret and reproduce observed actions under the guidance of a human user.

In order to perform a manipulation action, the robot also needs to learn what tool to grasp and on what object to perform the action. Our system applies CNN based recognition modules to recognize the objects and tools in the video. Then, given the beliefs of the tool and object (from the output of the recognition), our system predicts the most likely action using language, by mining a large corpus using a technique similar to [14]. Putting everything together, the output from the lower level visual perception system is in the form of (LeftHand GraspType1 Object1 Action RightHand GraspType2 Object2). We will refer to this septet of quantities as visual sentence.

At the higher level of representation, we generate a symbolic command sequence. [13] proposed a context-free grammar and related operations to parse manipulation actions. However, their system only processed RGBD data from a controlled lab environment. Furthermore, they did not consider the grasping type in the grammar. This work extends [13] by modeling manipulation actions using a

}

%% file: 2_rw.tex
\section{Related Work} \label{Sec:rw}

\textbf{Learning from Demonstration} LfD techniques are widely used in the robotics community to teach the robots new skills based on human demonstrations~\citep{Brenna2009}. Two popular LfD approaches are kinesthetic teaching~\citep{Pastor11, Akgun2012} and sensor-based demonstration~\citep{Calinon09, kruger2010learning}. Recently, ~\cite{Koenemann2014} introduced a method to transfer complex whole-body human motions to a humanoid robot in real-time. ~\cite{Welschehold2016} proposed to transform human demonstrations into hand-object trajectories in order to adapt to robotic manipulation tasks. The advantage of LfD approaches is their abilities to let the robots accurately repeat human motions, however, it is difficult to expand LfD techniques to a large number of tasks since the training process is usually designed for a specific task or needs training data from the real robotic systems~\citep{Akgun2012}.

\textbf{Action Representation} From a computer vision point of view, ~\cite{Aksoy2016} represented the continuous human actions as ``semantic event chains" and solved the problem as an activity detection task. ~\cite{Yang2015} proposed to learn manipulation actions from unconstrained videos using CNN and grammar based parser. However, they need an explicit representation of both the objects and grasping types to generate command sentences. ~\cite{lee2013syntactic} captured the probabilistic activity grammars from the data for imitation learning. ~\cite{Ramirez2015} extracted semantic representations from human activities in order to transfer skills to the robot. Recently, ~\cite{Plappert17} introduced a method to learn bidirectional mapping between human motion and natural language with RNN. In this paper, we propose to directly learn the commands from the demonstration videos without any prior knowledge. Our method takes advantage of CNN to extract robust features, and RNN to model the sequences, while being easily adapted to any human activity.

\textbf{Command Understanding} Currently, commands are widely used to communicate and control real robotic systems. However, they are usually carefully programmed for each task. This limitation means programming is tedious if there are many tasks. To automatically allow the robot to understand the commands, ~\cite{Tellex2011} formed this problem as a probabilistic graphical model based on the semantic structure of the input command. Similarly, ~\cite{Guadarrama13} introduced a semantic parser that used both natural commands and visual concepts to let the robot execute the task. While we retain the concepts of~\citep{Tellex2011, Guadarrama13}, the main difference in our approach is that we directly use the grammar-free commands from the translation module. This allows us to use a simple similarity measure to map each word in the generated command to the real command that can be used on the real robot.

\textbf{Video Captioning} With the rise of deep learning, the video captioning problem becomes more feasible to tackle. ~\cite{Donahue2014} introduced a first deep learning framework that can interpret an input video to a sentence. The visual features were first extracted from the video frames then fed into a LSTM network to generate the video captions. Recently, ~\cite{Haonan2016} introduced a new hierarchical RNN to generate one or multiple sentences to describe a video. ~\cite{Venugopalan2016} proposed a sequence-to-sequence model to generate the video captions from both RGB and optical flow images. The authors in~\citep{yao2015describing,Ramanishka2017cvpr} investigated the use of visual attention mechanism for this problem. Similarly, ~\cite{pan17_semantic_attributes} proposed to learn semantic attributes from the image then incorporated this information into a LSTM network. In this work, we cast the problem of translating videos to commands as a video captioning task to build on the strong state the art in computer vision. However, unlike~\citep{yao2015describing,Ramanishka2017cvpr} that explore the visual attention mechanism to improve the result, we focus on the fine-grained actions in the video. Our hypothesis is based on the fact that the fine-grained action is the key information in the video-to-command task, hence plays a significant contribution to the results.

\textbf{Reinforcement/Meta Learning} Recently, reinforcement learning and meta-learning techniques are also widely used to solve the problem of learning from human demonstrations. ~\cite{Rhinehart17_activity} proposed a method to predict the outcome of the human demonstrations from a scene using inverse reinforcement learning. ~\cite{Stadie17} learned a reward function based on human demonstrations of a given task in order to allow the robots to execute some trials, then maximize the reward to reach the desired outcome. With a different viewpoint, ~\cite{Finn17_oneshot_learning} proposed an one-shot visual imitation learning method to let the robot perform the tasks from just one single demonstration. More recently, ~\cite{yu2018one} presented a new approach for one-shot learning by fusing human and robot demonstration data to build up prior knowledge through meta-learning. The main advantage of meta-learning approach is it allows the robot to replicate human actions from just one or few demonstrations, however handling domain shift and the dependence on the data from the real robotic system are still the major challenges in this approach.

\junk{

Recently, the authors in~\citep{Aksoy2017} introduced an unsupervised method to link visual features to textual descriptions in long manipulation tasks.

  Furthermore, we use the output of the deep network as the input command to control a full-size humanoid robot, allowing it to perform different manipulation tasks.\\

Recently, the authors in~\cite{Matthias17} introduced a method to solve the problem of linking human whole-body motion and natural language as a captioning task.

, while avoid using the complicated natural language parser
This approach, however, is not straightforward to apply to real robotic applications since the complicated of human spoken language.

This work formed the problem as an activity detection task and did not be applied to robotic applications.

A drawback of this approach ...

The action understanding problem has been extensively studied in robotics and computer vision over the last few years. 

Our work differs in that we are concerned with completing a known task, and focus on when to ask questions as opposed to what type of question to ask

}

%% file: 3_rnn.tex
\section{Translating Videos to Commands} \label{Sec:rnn}
We start by formulating the problem and briefly describing three basic components (recurrent neural networks, visual feature extraction, and word embedding) that are used in our design. We then present in details the architecture of our video-to-command network (V2CNet) that simultaneously translates the input videos to robotic commands and classifies the fine-grained actions.

\subsection{Problem Formulation}
In order to effectively generate both the output command and understand the fine-grained action in the video, we cast the problem of translating videos to commands as a video captioning task, and use a TCN network to classify the actions. Both tasks use the same input feature and are jointly trained using a single multi-task loss function. In particular, the input video is considered as a list of frames, presented by a sequence of features  $\mathbf{X} = (\mathbf{x}_1, \mathbf{x}_2, ..., \mathbf{x}_n)$ from each frame. The output command is presented as a sequence of word vectors $\mathbf{Y}^w=(\mathbf{y}_1^w, \mathbf{y}_2^w, ..., \mathbf{y}_m^w)$, in which each vector $\mathbf{y}^w$ represents one word in the dictionary $D$. Each input video also has an action groundtruth label, represented as an one-hot vector $\mathbf{Y}^a \in {\mathbb{R}^{|C|}}$, where $C$ is the number of action classes, to indicate the index of the truth action class. Our goal is to simultaneously find for each sequence feature $\mathbf{X}_i$ its most probable command $\mathbf{Y}_i^w$, and its best suitable action class $\mathbf{Y}_i^a$. In practice, the number of video frames $n$ is usually higher than the number of words $m$. To make the problem more suitable for robotic applications, we use a dataset that contains mainly human demonstrations and assume that the command is in grammar-free format.

\subsection{Background}
\subsubsection{Recurrent Neural Networks\\}

\textbf{ LSTM} LSTM is a popular RNN since it can effectively model the long-term dependencies from the input data through its gating mechanism. The LSTM network takes an input $\mathbf{x}_t$ at each time step $t$, and computes the hidden state $\mathbf{h}_t$ and the memory cell state $\mathbf{c}_t$ as follows:

\begin{equation}
\label{Eq_LSTM} 
\begin{aligned} 
{\mathbf{i}_t} &= \sigma ({\mathbf{W}_{xi}}{\mathbf{x}_t} + {\mathbf{W}_{hi}}{\mathbf{h}_{t - 1}} + {\mathbf{b}_i})\\
{\mathbf{f}_t} &= \sigma ({\mathbf{W}_{xf}}{\mathbf{x}_t} + {\mathbf{W}_{hf}}{\mathbf{h}_{t - 1}} + {\mathbf{b}_f})\\
{\mathbf{o}_t} &= \sigma ({\mathbf{W}_{xo}}{\mathbf{x}_t} + {\mathbf{W}_{ho}}{\mathbf{h}_{t - 1}} + {\mathbf{b}_o})\\
{\mathbf{g}_t} &= \phi ({\mathbf{W}_{xg}}{\mathbf{x}_t} + {\mathbf{W}_{hg}}{h_{t - 1}} + {\mathbf{b}_g})\\
{\mathbf{c}_t} &= {\mathbf{f}_t} \odot {\mathbf{c}_{t - 1}} + {\mathbf{i}_t} \odot {\mathbf{g}_t}\\
{\mathbf{h}_t} &= {\mathbf{o}_t} \odot \phi ({\mathbf{c}_t})
\end{aligned}
\end{equation}
where $\odot$ represents element-wise multiplication, the function $\sigma: \mathbb{R} \mapsto [0,1], \sigma (x) = \frac{1}{{1 + {e^{ - x}}}}$ is the sigmod non-linearity, and $\phi: \mathbb{R} \mapsto [ - 1,1], \phi (x) = \frac{{{e^x} - {e^{ - x}}}}{{{e^x} + {e^{ - x}}}}$ is the hyperbolic tangent non-linearity. The parameters $\mathbf{W}$ and $\mathbf{b}$ are trainable weight and bias of the LSTM network. With this gating mechanism, the LSTM network can remember or forget information for long periods of time, while is still robust against the vanishing gradient problem. In practice, the LSTM network is straightforward to train end-to-end and is widely used in many problems~\citep{Donahue2014, nguyen2017_event_pose, Ramanishka2017cvpr, nguyen2018objectcaptioning}.

\textbf{GRU} Another popular RNN is Gated Recurrent Unit (GRU)~\citep{Cho14_GRU}. The main advantage of the GRU network is that it uses less computational resources in comparison with the LSTM network, while achieves competitive performance. Unlike the standard LSTM cell, in a GRU cell, the update gate controls both the input and forget gates, while the reset gate is applied before the nonlinear transformation as follows:

\begin{figure*}[!htbp] 

    \centering
	\includegraphics[scale=0.31]{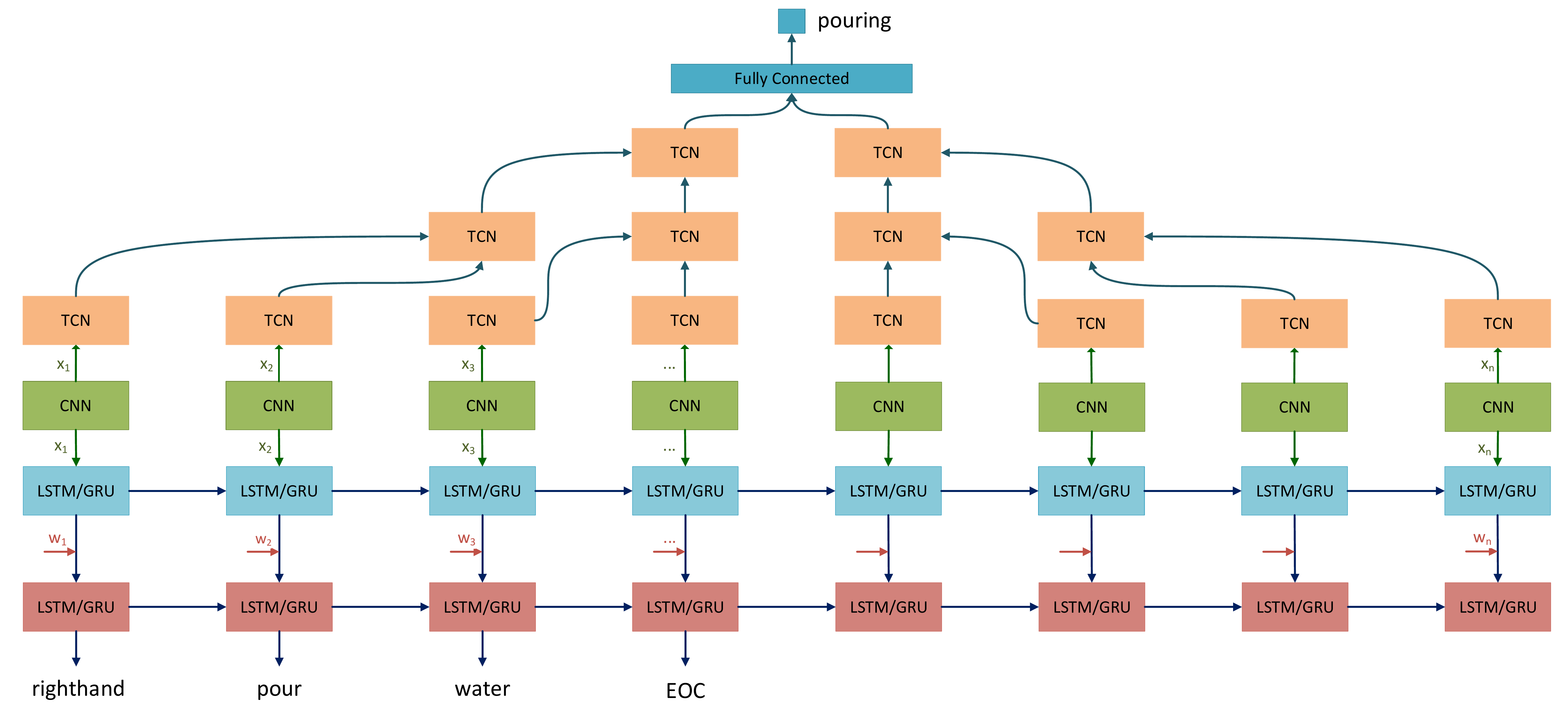} 
    \vspace{-2.0ex}
    \caption{A detailed illustration of our V2CNet architecture. The network is composed of two branches: a classification branch and a translation branch. The input for both branches is the visual features extracted by CNN. The classification branch uses the input features to learn the action through a TCN network. The translation branch has the encoder-decoder architecture with two LSTM/GRU layers. The first LSTM/GRU layer is used to encode the visual features, then the input words are combined with the output of the first LSTM/GRU layer and fed into the second LSTM/GRU layer in order to sequentially generate the output words as a command.}
    \label{Fig:overview} 
\end{figure*}

\begin{equation}
\label{Eq_GRU} 
\begin{aligned} 
{\mathbf{r}_t} &= \sigma ({\mathbf{W}_{xr}}{\mathbf{x}_t} + {\mathbf{W}_{hr}}{\mathbf{h}_{t - 1}} + {\mathbf{b}_r})\\
{\mathbf{z}_t} &= \sigma ({\mathbf{W}_{xz}}{\mathbf{x}_t} + {\mathbf{W}_{hz}}{\mathbf{h}_{t - 1}} + {\mathbf{b}_z})\\
{\mathbf{{\tilde h}}_t} &= \phi ({\mathbf{W}_{xh}}{\mathbf{x}_t} + {\mathbf{W}_{hh}}({\mathbf{r}_t} \odot {\mathbf{h}_{t - 1}}) + {\mathbf{b}_h}\\
{\mathbf{h}_t} &= {\mathbf{z}_t} \odot {\mathbf{h}_{t - 1}} + (1 - {\mathbf{z}_t}) \odot {\mathbf{{\tilde h}}_t}
\end{aligned}
\end{equation}
where $\mathbf{r}_t$, $\mathbf{z}_t$, and $\mathbf{h}_t$ represent the reset, update, and hidden gate of the GRU cell, respectively.

\subsubsection{Command Embedding}
Since a command is a list of words, we have to represent each word as a vector for computation. There are two popular techniques for word representation: \textit{one-hot} encoding and \textit{word2vec}~\citep{Mikolov2013} embedding. Although the one-hot vector is high dimensional and sparse since its dimensionality grows linearly with the number of words in the vocabulary, it is straightforward to use this embedding in the video captioning task. In this work, we choose the one-hot encoding technique as our word representation since the number of words in our dictionary is relatively small. The one-hot vector $\mathbf{y}^w \in {\mathbb{R}^{|D|}}$ is a binary vector with only one non-zero entry indicating the index of the current word in the vocabulary. Formally, each value in the one-hot vector $\mathbf{y}^w$ is defined by:

 \begin{equation}
    \mathbf{y}^w(i)=
    \begin{cases}
      1, & \text{if}\ i=ind(\mathbf{y}^w) \\
      0, & \text{otherwise}
    \end{cases}
  \end{equation}
where $ind(\mathbf{y}^w)$ is the index of the current word in the dictionary $D$. In practice, we add an extra word \textsf{EOC} to the dictionary to indicate the end of command sentences.

\subsubsection{Visual Features}
As the standard practice in the video captioning task~\citep{Venugopalan2016, Ramanishka2017cvpr}, the visual feature from each video frame is first extracted offline then fed into the captioning module. This step is necessary since we usually have a lot of frames from the input videos. In practice, we first sample $n$ frames from each input video in order to extract deep features from the images. The frames are selected uniformly with the same interval if the video is too long. In case the video is too short and there are not enough $n$ frames, we create an artificial frame from the mean pixel values of the ImageNet dataset~\citep{Olga2015} and pad this frame at the end of the list until it reaches $n$ frames. We then use the state-of-the-art CNN to extract deep features from these input frames. Since the visual features provide the key information for the learning process, three popular CNN are used in our experiments: VGG16~\citep{SimonyanZ14}, Inception\_v3~\citep{Szegedy16_Inception}, and ResNet50~\citep{He2016}.

Specifically, for the VGG16 network, the features are extracted from its last fully connected $\texttt{fc2}$ layer. For the Inception\_v3 network, we extract the features from its $\texttt{pool\_3:0}$ tensor. Finally, we use the features from $\texttt{pool5}$ layer of the ResNet50 network. The dimension of the extracted features is $4096$, $2048$, $2048$, for the VGG16, Inception\_v3, and ResNet50 network, respectively. All these CNN are pretrained on ImageNet dataset for image classifications. We notice that the names of the layers we mention here are based on the Tensorflow~\citep{TensorFlow2015} implementation of these networks.

\subsection{Architecture}
\subsubsection{Overview}
To simultaneously generate the command sentence and classify the fine-grained actions, we design a deep network with two branches: a classification branch and a translation branch. The classification branch is a TCN network that handles the human action, while the translation branch is a RNN with encoder-decode architecture to encode the visual features and sequentially generate the output words as a command. Both branches share the input visual features from the video frames and are trained end-to-end together using a single multi-task loss function. Our intuition is that since the translation branch has to generate the command from a huge space of words, it may not effectively encode the fine-grain actions. Therefore, we integrate the classification branch which is trained directly on the smaller space of action classes to ease the learning process. In practice, the parameters of both branches are updated using the same gradient signal, hence the classification branch will encourage the translation branch to generate the correct fine-grained action. Fig.~\ref{Fig:overview} illustrates the details of our V2CNet network.

\subsubsection{Fine-grained Action Classification} We employ a TCN network in the classification branch to encode the temporal information of the video in order to classify the fine-grained actions. Unlike the popular 2D convolution that operates on the 2D feature map of the image, the TCN network performs the convolution across the time axis of the data. Since each video is represented as a sequence of frames and the fine-grained action is the key information for the learning process, using the TCN network to encode this information is a natural choice. Recent work showed that the TCN network can further improve the results in many tasks such as action segmentation and detection~\citep{sun2015_action, Colin17_temporal}.

The input of the classification branch is a set of visual features extracted from the video frames by a pretrained CNN. The output of this branch is a classification score that indicates the most probable action for the input video. In practice, we build a TCN with three convolutional layers to classify the fine-grained actions. After each convolutional layer, a non-linear activation layer (ReLU) is used, followed by a max pooling layer to gradually encode the temporal information of the input video. Here, we notice that instead of performing the max pooling operation in the 2D feature map, the max pooling of TCN network is performed across the time axis of the video. Finally, the output of the third convolutional layer is fed into a fully connected layer with $256$ neurons, then the last layer with only $1$ neuron is used to regress the classification score for the current video.

More formally, given the input feature vector $\mathbf{X}$, three convolutional layers $\mathbf{\Phi}_c$ of the TCN network are defined as follows to encode the temporal information across the time axis of the input demonstration video:

\begin{equation}
\label{eq_tcn_conv} 
\begin{aligned} 
{\mathbf{\Phi}_{c0}}({\mathbf{W}_{c0}},{\mathbf{b}_{c0}}) &= {\mathbf{W}_{c0}}\mathbf{X} + {\mathbf{b}_{c0}}  \\
{\mathbf{\Phi}_{c1}}({\mathbf{W}_{c1}},{\mathbf{b}_{c1}}) &= {\mathbf{W}_{c1}}(\text{MaxPool}(\text{ReLU}({\mathbf{\Phi} _{c0}})) + {\mathbf{b}_{c1}}  \\
{\mathbf{\Phi}_{c2}}({\mathbf{W}_{c2}},{\mathbf{b}_{c2}}) &= {\mathbf{W}_{c2}}(\text{MaxPool}(\text{ReLU}({\mathbf{\Phi} _{c1}})) + {\mathbf{b}_{c2}}  \\
\end{aligned}
\end{equation}
then the third convolutional layer ${\mathbf{\Phi}_{c2}}$ is fed into a fully connected layer $\mathbf{\Phi} _{f0}$ as follows:

\begin{equation}
\label{eq_tcn_fc} 
\begin{aligned} 
{\mathbf{\Phi} _{f0}} ({\mathbf{W}_{f0}},{\mathbf{b}_{f0}}) &= {\mathbf{W}_{f0}}({\text{ReLU} (\mathbf{\Phi} _{c2}})) + {\mathbf{b}_{f0}} \\
{\mathbf{\Phi} _{f1}} ({\mathbf{W}_{f1}},{\mathbf{b}_{f1}}) &= {\mathbf{W}_{f1}}{\mathbf{\Phi} _{f0}} + {\mathbf{b}_{f1}}  \\
\end{aligned}
\end{equation}
where $\mathbf{W}_c$, $\mathbf{b}_c$ and  $\mathbf{W}_f$, $\mathbf{b}_f$ are the weight and bias of the convolutional and fully connected layers. In practice, the filter parameters of three convolutional layers $\mathbf{\Phi}_{c0}$, $\mathbf{\Phi}_{c1}$, $\mathbf{\Phi}_{c2}$ are empirically set to $2048$, $1024$, and $512$, respectively. Note that, the ReLU activation is used in the first fully-connected $\mathbf{\Phi} _{f0}$ layer, while there is no activation in the last fully connected layer $\mathbf{\Phi}_{f1}$ since we want this layer outputs a probability of the classification score for each fine-grained action class.

\subsubsection{Command Generation} In parallel with the classification branch, we build a translation branch to generate the command sentence from the input video. The architecture of our translation branch is based on the encoder-decoder scheme~\citep{Venugopalan2016, Ramanishka2017cvpr}, which is adapted from the popular sequence to sequence model~\citep{Sutskever2014_Seq} in machine translation. Although recent approaches use the visual attention mechanism~\citep{yao2015describing, Ramanishka2017cvpr} or hierarchical RNN~\citep{Haonan2016} to improve the results, our translation branch solely relies on the neural architecture. However, since the translation branch is jointly trained with classification branch, it is encouraged to output the correct fine-grained action as learned by the classification branch. We experimentally show that by simultaneously training both branches, the translation accuracy is significantly improved over the state of the art. 

In particular, the translation branch also uses the visual features extracted by a pretrained CNN as in the classification branch. These features are encoded in the first RNN layer to create the encoder hidden state. The input words are then combined with the hidden state of the first RNN layer, and fed into the second RNN layer. This layer decodes its input and sequentially generates a list of words as the output command. More formally, given an input sequence of features $\mathbf{X} = (\mathbf{x}_1, \mathbf{x}_2, ..., \mathbf{x}_n)$, we want to estimate the conditional probability for an output command  $\mathbf{Y}^w=(\mathbf{y}_1^w, \mathbf{y}_2^w, ..., \mathbf{y}_m^w)$ as follows:
\begin{equation} \label{Eq_mainP1}
P(\mathbf{y}_1^w, ..., \mathbf{y}_m^w | \mathbf{x}_1, ..., \mathbf{x}_n) = \prod\limits_{i = 1}^m {P({\mathbf{y}_i^w}|{\mathbf{y}_{i - 1}^w}, ...,{\mathbf{y}_1^w}},\mathbf{X})
\end{equation}

Since we want a generative model that encodes a sequence of features and produces a sequence of words in order as a command, the LSTM/GRU is well suitable for this task. Another advantage of LSTM/GRU is that they can model the long-term dependencies in the input features and the output words. In practice, we use LSTM and GRU as our recurrent neural network, while the input visual features are extracted from the VGG16, Inception\_v3, and ResNet50 network, respectively.

In the encoding stage, the first LSTM/GRU layer converts the visual features $\mathbf{X} = (\mathbf{x}_1, \mathbf{x}_2, ..., \mathbf{x}_n)$ to a list of hidden state vectors $\mathbf{H}^e = (\mathbf{h}_1^e, \mathbf{h}_2^e, ..., \mathbf{h}_n^e)$ (using Equation~\ref{Eq_LSTM} for the LSTM network or Equation~\ref{Eq_GRU} for the GRU network). Unlike~\cite{venugopalan2014translating} which takes the average of all $n$ hidden state vectors to create a fixed-length vector, we directly use each hidden vector $\mathbf{h}_i^e$ as the input $\mathbf{x}_i^d$ for the second decoder layer. This allows the smooth transaction from the visual features to the output commands without worrying about the harsh average pooling operation, which can lead to the loss of temporal structure underlying the input video.

In the decoding stage, the second LSTM/GRU layer converts the list of hidden encoder vectors $\mathbf{H}^e$ into the sequence of hidden decoder vectors $\mathbf{H}^d$. The final list of predicted words $\mathbf{\hat{Y}}$ is achieved by applying a softmax layer on the output $\mathbf{H}^d$ of the LSTM/GRU decoder layer. In this way, the LSTM/GRU decoder layer sequentially generates a conditional probability distribution for each word of the output command given the encoded features representation and all the previously generated words. In practice, we preprocess the data so that the number of input words $m$ is equal to the number of input frames $n$. For the input video, this is done by uniformly sampling $n$ frames in the long video, or padding the extra frame if the video is too short. Since the number of words $m$ in the input commands is always smaller than $n$, we pad a special empty word to the list until we have $n$ words.

\subsubsection{Multi-Task Loss}
We train the V2CNet using a joint loss function for both the classification and translation branches as follows:
\begin{equation}
L = L_{cls} + L_{trans}
\end{equation}
where the $L_{cls}$ loss is defined in the classification branch for fine-grained action classification, and $L_{trans}$ is defined in the RNN branch for command generation.

Specially, $L_{cls}$ is the sigmoid cross entropy loss over the groundtruth action classes $C$, and is defined as follows:
\begin{equation}
{L_{cls}} =  - \sum\limits_{i = 1}^C {{\mathbf{y}_i^a}\log ({\hat{\mathbf{y}}_i^a})}
\end{equation}
where $\mathbf{y}^a$ is the groundtruth action label of the current input video, and $\mathbf{\hat y}^a$ is the predicted action output of the classification branch of the network. 

We follow the process in~\citep{Donahue2014} to compute the $L_{trans}$ loss for the translation branch. Let $\mathbf{z}_t$ denote the output of each cell at each time step $t$ in the second LSTM/GRU layer in the translation branch. This output is then passed through a linear prediction layer $\hat{\mathbf{y}}_t=\mathbf{W}_{z}\mathbf{z}_t+\mathbf{b}_z$, and the predicted distribution $P(\mathbf{y}_t)$ is computed by taking the softmax of $\hat{\mathbf{y}}_t$ as follows:

\begin{equation}\label{eq_loss_cmd1}
P(\mathbf{y}_t=\mathbf{w}|\mathbf{z}_t)=\frac{\text{exp}(\hat{\mathbf{y}}_{t,w})}{\sum_{w'\in{D}}\text{exp}(\hat{\mathbf{y}}_{t,w'})}
\end{equation}
where $\mathbf{W}_z$ and $\mathbf{b}_z$ are learned parameters, $\mathbf{w}$ is a word in the dictionary $D$. The $L_{trans}$ loss is then computed as follows:

\begin{equation}\label{eq_loss_cmd2}
{L_{trans}} = {\frac{1}{m}\sum\limits_{t = 1}^m {{P}({{\bf{y}}_t} = {\bf{w}}|{{\bf{z}}_t})} }
\end{equation}

Intuitively, Equation~\ref{eq_loss_cmd2} computes the softmax loss at each word of the current output command given all the previous generated words.

\subsubsection{Training and Inference \\}

\textbf{     } During the training phase, the classification branch uses the visual features to learn the fine-grained actions using the TCN network. In parallel with the classification branch, the translation branch also receives the input via its first LSTM/GRU layer. In particular, at each time step $t$, the input feature $\mathbf{x}_t$ is fed to an LSTM/GRU cell in the first LSTM/GRU layer along with the previous hidden state $\mathbf{h}_{t-1}^e$ to produce the current hidden state $\mathbf{h}_t^e$. After all the input features are exhausted, the word embedding and the hidden states of the first LSTM/GRU layer are fed to the second LSTM/GRU layer. This layer converts the inputs into a sequence of words by maximizing the log-likelihood of the predicted word. This decoding process is performed sequentially for each word until the network generates the \textsf{EOC} token. Since both the classification and translation branches are jointly trained using a single loss function, the weight and bias parameters of both branches are updated using the same gradient signal through backpropagation. 

During the inference phase, the input for the network is only the visual features of the testing video. The classification branch uses these visual features to generate the probabilities for all action classes. The final action class is chosen from the class with the highest classification score. Similarly, the visual features are fed into two LSTM/GRU layers to sequentially generate the output words as the command. Note that, unlike the training process, during inference the input for the second LSTM/GRU layer in the translation branch is only the hidden state of the first LSTM/GRU layer. The final command sentence is composed of the first generated word and the last word before the \textsf{EOC} token.

\junk{

{L_{cls}} = {\log (1 + \text{exp}( - \mathbf{y}^a \mathbf{\hat y}^a)}

   training both branches, we encourage translation branch to generate the output command with the correct human action as generated by the classification branch.

  which makes it   While translation branch is responsible for generating the command, the classification branch can be considered as a non visual attention mechanism since it enables the network to output the correct human action, which is the key information in this task. By

Both the classification and translation branches are trained end-to-end with Adam optimizer~\citep{kingma2014adam} with the learning rate of $0.0001$.

 Based on the input data characteristics, the captioning branch smoothly encodes the input visual features and generates the output commands, achieving a fair improvement over the state of the art without using any additional modules.

 The core of an LSTM network is a memory cell $\mathbf{c}$ which has the gate mechanism to encode the knowledge of the previous inputs at every time step.

In practice, the LSTM network is straightforward to train end-to-end and can handle inputs with different lengths using the padding techniques.


\begin{equation}
\Phi (\mathbf{X},\theta ) = {\mathbf{W}_2}\text{ReLU}({\mathbf{W}_1}\text{ReLU}({\mathbf{W}_0}\mathbf{X} + {\mathbf{b}_0}) + {\mathbf{b}_1}) + {\mathbf{b}_2}
\end{equation}
where the parameters of the network $\theta$ is defined by $\phi  = \{ {W_0} \in {R^{30 \times 2048}},{W_1} \in {R^{15 \times 1024}},{W_2} \in {R^{8 \times 512}},{b_0} \in {R^{30}},{b_1} \in {R^{15}},{b_2} \in {R^8}$.

In this section we define two TCNs, each of which have the following properties: (1) computations are performed layer-wise, meaning every time-step is updated simultane- ously, instead of updating sequentially per-frame (2) con- volutions are computed across time, and (3) predictions at each frame are a function of a fixed-length period of time, which is referred to as the receptive field

This layer models the probability distribution of the next word over the word space as follow:
  
The key different in our architecture in comparision with~\cite{venugopalan2014translating} is we do not do average pooling when feeding frame features to LSTM/GRU, instead each feature is fed directly to LSTM/GRU. If the number of frame is less than the number of LSTM step, we pad the missing frames with the mean image from ImageNet dataset. TODO: Compare more.

In the first several time steps, the top LSTM layer (colored red in Figure 2) receives a se- quence of frames and encodes them while the secondLSTM layer receives the hidden representation (ht) and concate- nates it with null padded input words (zeros), which it then encodes. There is no loss during this stage when the LSTMs are encoding. After all the frames in the video clip are ex- hausted, the second LSTM layer is fed the beginning-of- sentence (<BOS>) tag, which prompts it to start decoding its current hidden representation into a sequence of words. While training in the decoding stage, the model maximizes for the log-likelihood of the predicted output sentence given the hidden representation of the visual frame sequence, and the previouswords it has seen. From Equation 3 for a model with parameters θ and output sequence Y = (y1, . . . , ym), this is formulated as

During training the visual feature, sentence pair (V,S) is provided to the model, which then optimizes the log-likelihood (Equation 1) over the entire training dataset using stochastic gradient descent. At each time step, the input xt is fed to the LSTM along with the previous time step’s hidden state ht−1 and the LSTM emits the next hidden state vector ht (and a word). For the first layer of the LSTM xt is the con- catenation of the visual feature vector and the pre- vious encoded word, the ground truth word during training and the predicted word during test time). For the second layer of the LSTM xt is zt of the first layer. Accordingly, inference must also be performed sequentially in the order h1 = fW(x1, 0), h2 = fW(x2,h1), until the model emits the end- of-sentence (EOS) token at the final step T. In our model the output (ht = zt) of the second layer LSTM unit is used to obtain the emitted word

 It updates its hid- den state every time a new word arrives, and encodes the sentence semantics in a compact form up to the words that have been fed in.

 decoder is conditioned step by step on the first t words of the caption and on the corresponding video descriptor, and is tr111ained to produce the next word of the caption. The objective function which we optimize is the log-likelihood of correct words over the sequence

In practice, (THEORY) - PRACTICE These features are then fed to a RNN (LSTM or GRU) to encode them to a fixed length vector $\mathbf{z}$. This process is done by extracting the hidden state vector $h_i^e$ from LSTM/GRU. The middle state is computed by taking the average of all $m$ feature descriptors:

\begin{equation}
z = \frac{1}{m}\sum\limits_{i = 1}^m {{x_i}}
\end{equation}

The decoder converts the encoded vector $z$ into output sequence of words $y_t$, $t \in {1, . . . ,n}$. In particular,

Specifically, for the VGG16 net, the feature map is extracted from its last fully connected layer (i.e. $\textit{fc2} \in {\mathbb{R}^{4096}}$). For the Inception\_v3 network, we extract the features from its $\textit{pool\_3:0} \in {\mathbb{R}^{2048}}$ tensor. Finally, we use the features from $\textit{pool5} \in {\mathbb{R}^{2048}}$ layer of ResNet50.

During the training phase, the frame features and  groundtruth one-hot vectors are used to reproduce these vectors. In the test time, the list of one-hot vectors is generated based on just the input features, which represents the output sentence.

We apply our approach to both still image and video description scenarios, adapting a popular encoder-decoder model for video captioning [22] as our base model.

In this section, we first briefly review two popular RNN that we use in our experiments: Long Short-Term Memory (LSTM)~\cite{Hochreiter97_LSTM} and a variation of the LSTM called Gated Recurrent Units (GRU)~\cite{Cho14_GRU}. We then describe our approach to directly translate videos to robot commands. Fig.~\ref{Fig:overview} shows an overview of our approach.

Commands: list of words (in grammar free format)
Video: list of frames represent a short human action. Although we can describe anything, in this paper, we only focus on manipulation actions of human. More formally, let: 

\begin{equation}
 F = ({f_1},{f_2},...,{f_n})
 \end{equation} 
represent a list of features extracted from the video, and:
\begin{equation}
 C = ({c_1},{c_2},...,{c_m})
 \end{equation} 
 
represent a list of words in a command.

that We first briefly describe the Long-Short Term Memory (LSTM) network, which has shown great success to sequential problems such as machine translation~\cite{s}, image captioning~\cite{•}

Recently, Recurrent Neural Networks (RNNs) have received a lot of attention and achieved impressive results in various applications, including speech recognition [25], image captioning [26] and video description [27]. RNNs have also been applied to facial landmark detection. Very recently, Peng et al. [28] propose a recurrent encoder-decoder network for video-based sequential facial landmark detection. They use recurrent networks to align a sequence of the same face in video frames, by exploiting recurrent learning in spatial and temporal dimensions.

In this section, we propose an approach to translate the input videos to robot commands. Our approach is based on deep RNN models (LSTM and GRU), which have been successfully used in various problems such as machine translation~\cite{Ilya2014}, image captioning~\cite{Donahue2014}, video description~\cite{Venugopalan2016}.

These one-hot vectors are then projected into an embedding space with dimension de by multiplication Weyt with a learned parameter matrix We ∈ Rde×K. The result of a matrix- vector multiplication with a one-hot vector is the column of the matrix corresponding to the index of the single non- zero component of the one-hot vector. We can therefore be thought of as a “lookup table,” mapping each of the K words in the vocabulary to a de-dimensional vector.

plus one additional entry for the <BOS> (beginning of sequence) token which is always taken as y0, the vector $y \int RK$ with a single non-zero component yi = 1 denoting the ith word in the vocabulary. 
a binary vector with exactly one non-zero entry at the position indicating the index of the word in the vocabulary

In addition, the proposed recurrent network in this paper contains multiple Long-Short Term Memory (LSTM) components. For self-containness, we give a brief introduction to LSTM. LSTM is one type of the RNNs, which is attractive because it is explicitly designed to remember information for long periods of time. LSTM takes xt, ht−1 and ct−1 as inputs, ht and ct as outputs

where σ(x) = (1 + e−x)−1 is the sigmoid function. The outputs of the sigmoid functions are in the range of [0, 1], here the smaller values indicate “more probability to forget the previous information” and the larger values indicate “more probability to remember the information”. tanh(x)is the tangent non-linearity function and its outputs are in the range [−1, 1]. [x;h] represents the concatenation of x and h. LSTM has four gates, a forget gate ft, an input gate it, an output gate ot and an input modulation gate gt. ht and ct are the outputs. ht is the hidden unit. ct is the memory cell that can be learned to control whether the previous information should be forgotten or remembered.

At time step t, the input to the bottom-most LSTM is the embedded word from the previous time step yt−1. Input words are encoded as one-hot vectors: vectors $y \int RK$ with a single non-zero component yi = 1 denoting the ith word in the vocabulary, where K is the number of words in the vocabulary, plus one additional entry for the <BOS> (beginning of sequence) token which is always taken as y0, the “previous word” at the first time step (t = 1). These one-hot vectors are then projected into an embedding space with dimension de by multiplication Weyt with a learned parameter matrix We ∈ Rde×K. The result of a matrix- vector multiplication with a one-hot vector is the column of the matrix corresponding to the index of the single non- zero component of the one-hot vector. We can therefore be thought of as a “lookup table,” mapping each of the K words in the vocabulary to a de-dimensional vector.
}

%% file: 4_exp.tex
\section{Experiments} \label{Sec:exp}
\subsection{Dataset}

Recently, many datasets have been proposed in the video captioning field~\citep{Xu16_MSR_Dataset}. However, these datasets only provide general descriptions of the video and there is no detailed understanding of the action. The groundtruth captions are also written using natural language sentences which can not be used directly in robotic applications. Motivated by these limitations, we introduce a new \textit{video-to-command} (IIT-V2C) dataset which focuses on \textit{fine-grained} action understanding~\citep{lea2016learning}. Our goal is to create a new large-scale dataset that provides fine-grained understanding of human actions in a grammar-free format. This is more suitable for robotic applications and can be used with deep learning methods.

\begin{figure}[ht]
\centering
\footnotesize

  \stackunder[2pt]{\includegraphics[width=0.99\linewidth, height=0.18\linewidth]{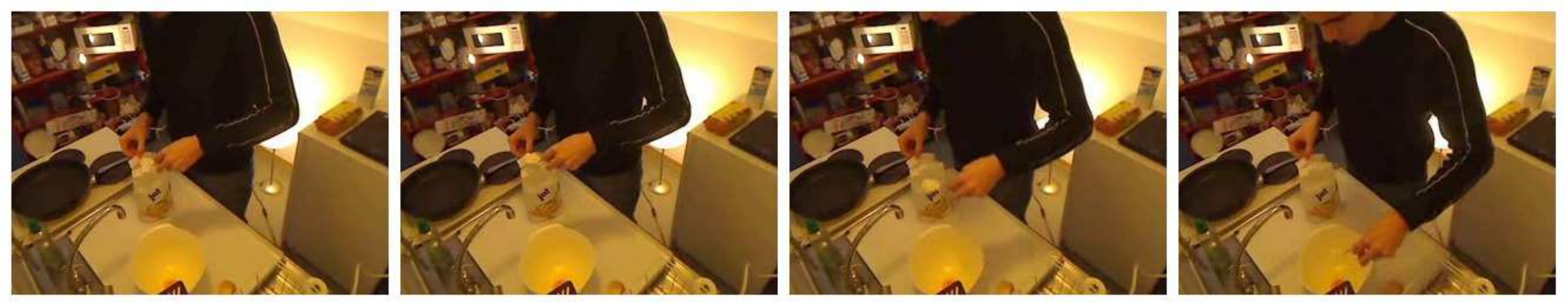}}  				
  				  {\datasetCaption {lefthand transfer powder to bowl} {transferring}}
  \stackunder[2pt]{\includegraphics[width=0.99\linewidth, height=0.18\linewidth]{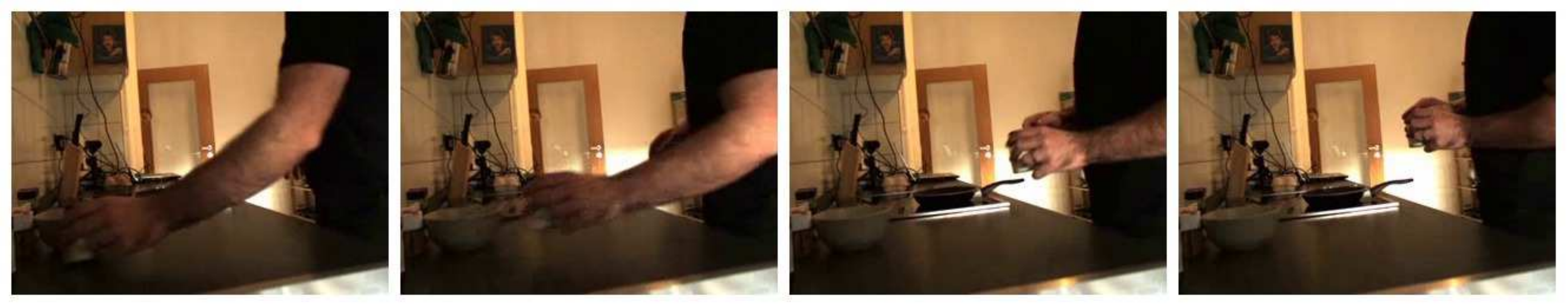}}  
  				  {\datasetCaption {lefthand carry salt box} {carrying}}

  \stackunder[2pt]{\includegraphics[width=0.99\linewidth, height=0.18\linewidth]{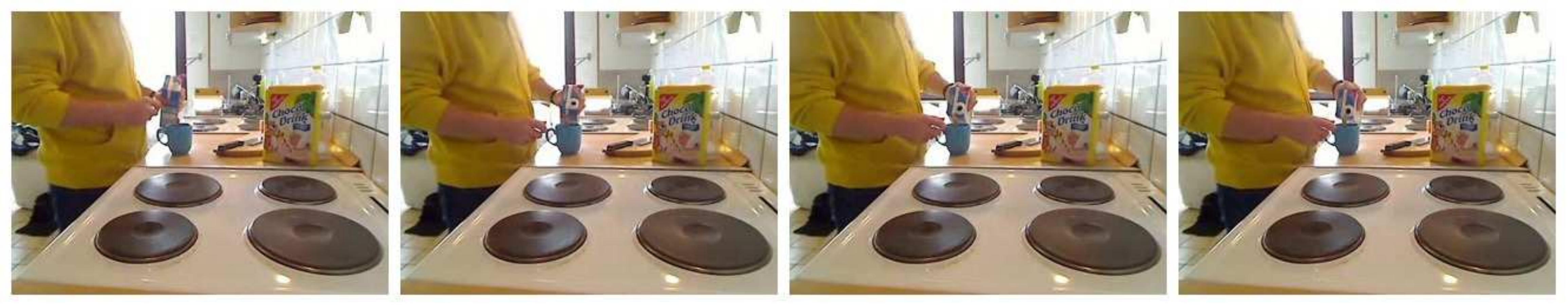}}  
    			  {\datasetCaption {righthand pour milk} {pouring} }
  
  \stackunder[2pt]{\includegraphics[width=0.99\linewidth, height=0.18\linewidth]{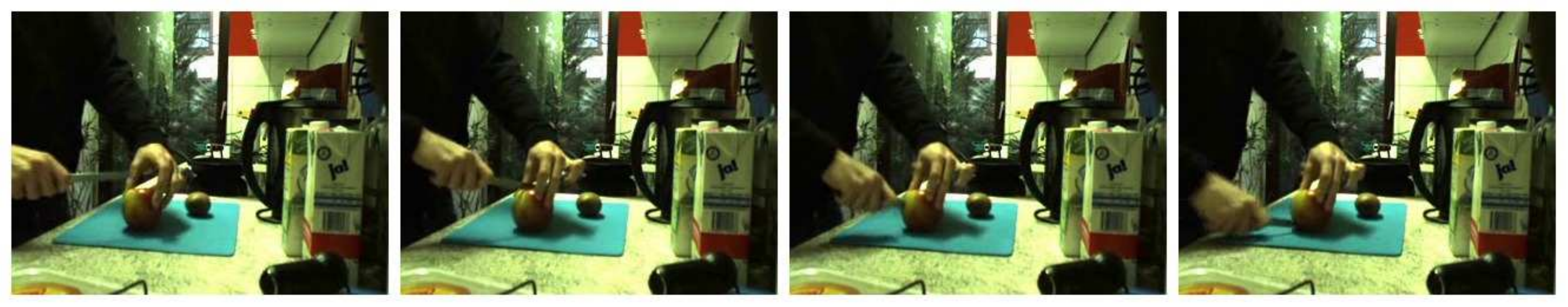}}  
    			  {\datasetCaption {bothhand cut apple} {cutting} }
  				  
\vspace{-0.0ex}
\caption{Example of human demonstration clips and labeled groundtruths for the command sentence and fine-grained action class in our IIT-V2C dataset. The clips were recorded from challenging viewpoints with different lighting conditions.}
\label{fig_exp_figure} 
\end{figure}

\begin{figure*}[ht]
  \centering
 	\includegraphics[scale=0.32]{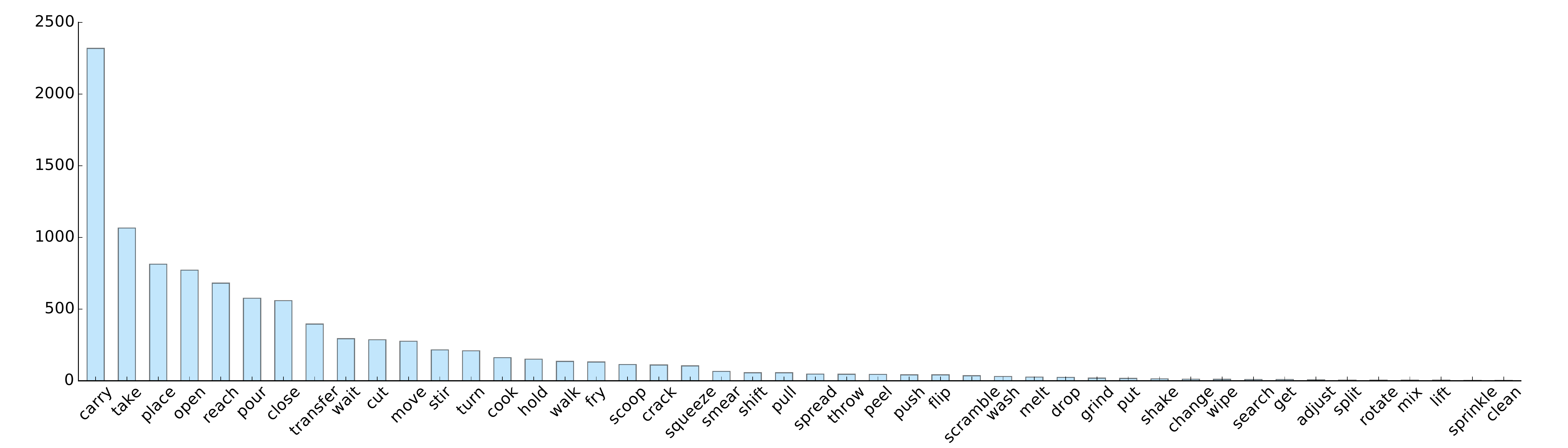}
 \vspace{-1.0ex}
 \caption{The distribution of the fine-grained action classes in our IIT-AFF dataset.}
 \label{fig_act_distribution}
\end{figure*}

\textbf{Video annotation} 
Since our main purpose in this work is to develop a framework that can be used by real robots for manipulation tasks, we are only interested in the videos that have human demonstrations. To this end, the raw videos in the Breakfast dataset~\citep{Kuehne14_BF_Dataset} are best suited to our purpose since they were originally designed for activity recognition. We only reuse the raw videos from the Breakfast dataset and manually segment each video into short clips in a fine granularity level. Each short clip is then annotated with a \textit{command sentence} in grammar-free that describes the current human action. For each command sentence, we use the Stanford POS Tagger~\citep{Toutanova_tagger} to automatically extract the verb from the command. This extracted verb is then considered as the fine-grained \textit{action groundtruth} for the current video.

\newcolumntype{g}{>{\columncolor{Gray}}c}
\begin{table*}[!ht]
\centering\ra{1.2}
\caption{The translation results on IIT-V2C dataset.}
\renewcommand\tabcolsep{2.5pt}
\label{tb_result_v2c}
\hspace{2ex}

\begin{tabular}{@{}r|c|c|cccccccc@{}}  
\toprule 					 &
\small RNN  	& 
\small Feature  & 

\small Bleu\_1  & 
\small Bleu\_2  & 
\small Bleu\_3  &
\small Bleu\_4  & 
\small METEOR  &
\small ROUGE\_L &
\small CIDEr \\

\midrule

S2VT~\citep{Venugopalan2016} 						&LSTM & VGG16, AlexNet		& 0.383   & 0.265   & 0.201	& 0.159	& 0.183    & 0.382   & 1.431     \\
S2VT~\citep{Venugopalan2016} 						&LSTM & ResNet50, AlexNet	& 0.397   & 0.280   & 0.219	& 0.177	& 0.196    & 0.401   & 1.560     \\
SGC~\citep{Ramanishka2017cvpr}						&LSTM & Inception   	& 0.370   & 0.256   & 0.198	& 0.161	& 0.179    & 0.371   & 1.422     \\
SCN~\citep{SCN_CVPR2017}							&LSTM & ResNet50, C3D  	& 0.398   & 0.281   & 0.219	& 0.190	& 0.195    & 0.399   & 1.561     \\

\cline{1-10}
\multirow{6}{*}{EDNet~\citep{Nguyen_V2C_ICRA18}} & \multirow{3}{*}{LSTM} 	&VGG16 		& 0.372   & 0.255  	& 0.193	& 0.159	& 0.180    & 0.375   & 1.395     \\
															  &	&Inception	& 0.400   & 0.286   & 0.221	& 0.178	& 0.194    & 0.402   & 1.594     \\   
												 			  &	&ResNet50	& 0.398   & 0.279   & 0.215	& 0.174	& 0.193    & 0.398   & 1.550     \\
									  \cline{2-10}
									  & \multirow{3}{*}{GRU}
																&VGG16 		& 0.350   & 0.233   & 0.173	& 0.137	& 0.168    & 0.351   & 1.255     \\
														 	  &	&Inception	& 0.391   & 0.281  	& 0.222	& 0.188	& 0.190    & 0.398   & 1.588     \\
													 		  &	&ResNet50	& 0.398   & 0.284   & 0.220	& 0.183	& 0.193    & 0.399   & 1.567     \\
\cline{1-10}

\multirow{6}{*}{V2CNet (ours)} & \multirow{3}{*}{LSTM}	& VGG16 	& 0.391   & 0.275   & 0.212 & 0.174 & 0.189    & 0.393   & 1.528     \\
					   									  	&   & Inception & 0.401   & 0.289   & 0.227 & 0.190 & 0.196    & 0.403   & 1.643     \\
					   									  	&   & ResNet50 	&\textbf{0.406}   & \textbf{0.293}   & \textbf{0.233} & \textbf{0.199} & \textbf{0.198}    & \textbf{0.408}   & \textbf{1.656}	 \\
					   			   \cline{2-10}
					   			   & \multirow{3}{*}{GRU} 		& VGG16 	& 0.389   & 0.267   & 0.208 & 0.172 & 0.186    & 0.387   & 1.462     \\
					  									  	&   & Inception & 0.402   & 0.285   & 0.224 & 0.189 & 0.196    & 0.405   & 1.618     \\
					  									  	&   & ResNet50  & 0.403   & 0.288   & 0.226 & 0.191 & 0.196    & 0.403   & 1.596     \\

\bottomrule
\end{tabular}
\end{table*}

\textbf{Dataset statistics} Overall, we reuse $419$ videos from the Breakfast dataset. These videos were captured when humans performed cooking tasks in different kitchens, then encoded with $15$ frames per second. The resolution of demonstration videos is $320 \times 240$. We segment each video (approximately $2-3$ minutes long) into around $10-50$ short clips (approximately $1-15$ seconds long), resulting in $11,000$ unique short videos. Each short video has a single command sentence that describes human actions. From the groundtruth command sentence, we extract the verb as the action class for each video, resulting in an action set with $46$ classes (e.g., cutting, pouring, etc.). Fig.~\ref{fig_exp_figure} shows some example frames of groudtruth command sentences and action classes in our new dataset. In Fig.~\ref{fig_act_distribution}, the distribution of the fine-grained action classes is also presented. Although our new-form dataset is characterized by its grammar-free property for the convenience in robotic applications, it can easily be adapted to classical video captioning task by adding the full natural sentences as the new groundtruth for each video.

\subsection{Evaluation, Baseline, and Implementation}
\textbf{Evaluation Metric} We use the standard evaluation metrics in the video captioning field~\citep{Xu16_MSR_Dataset} (Bleu, METEOR, ROUGE-L, and CIDEr) to report the translation results of our V2CNet. By using the same evaluation metrics, we can directly compare our results with the recent state of the art in the video captioning field.

\textbf{Baseline} The translation results of our V2CNet are compared with the following state of the art in the field of video captioning: S2VT~\citep{Venugopalan2016}, SGC~\citep{Ramanishka2017cvpr}, and SCN~\citep{SCN_CVPR2017}. In S2VT, the authors used the encoder-decoder architecture with LSTM to encode the visual features from RGB images (extracted by ResNet50 or VGG16) and optical flow images (extracted by AlexNet~\citep{krizhevsky2012imagenet}). In SGC, the authors also used the encoder-decoder architecture and LSTM, however, this work integrated a saliency guided method as the visual attention mechanism, while the visual features are from the Inception network. The authors in SCN~\citep{SCN_CVPR2017} first combined the visual features from ResNet and temporal features from C3D~\citep{Tran15_C3D} network, then extracted semantic concepts (i.e., tags) from the video frames. All these features were learned in a LSTM as a semantic recurrent neural network. Finally, we also compare the V2CNet results with our early work (EDNet~\citep{Nguyen_V2C_ICRA18}). The key difference between EDNet and V2CNet is the EDNet does not use the TCN network to jointly train the fine-grained action classification branch and the translation branch as in the V2CNet. For all methods, we use the code provided by the authors of the associated papers for the fair comparison.

\textbf{Implementation} We use $512$ hidden units in both LSTM and GRU in our implementation. The first hidden state of LSTM/GRU is initialized uniformly in $[-0.1, 0.1]$. We set the number of frames for each input video at $n=30$. Subsequently, we consider each command has maximum $30$ words. If there are not enough $30$ frames/words in the input video/command, we pad the mean frame/empty word at the end of the list until it reaches $30$. The mean frame is composed of pixels with the same mean RGB value from the ImageNet dataset (i.e., $(104, 117, 124)$). We use $70\%$ of the IIT-V2C dataset for training and the remaining $30\%$ for testing. During the training phase, we only accumulate the softmax losses of the real words to the total loss, while the losses from the empty words are ignored. We train all the variations of V2CNet for $300$ epochs using Adam optimizer~\citep{kingma2014adam} with a learning rate of $0.0001$, and the batch size is empirically set to $16$. The training time for each variation of the V2CNet is around $8$ hours on an NVIDIA Titan X GPU.

\subsection{Translation Results}

Table~\ref{tb_result_v2c} summarizes the translation results on the IIT-V2C dataset. This table clearly shows that our V2CNet with ResNet50 feature achieves the highest performance in all metrics: Blue\_1, Blue\_2, Blue\_3, Blue\_4, METEOR, ROUGE\_L, and CIDEr. Our proposed V2CNet also outperforms recent the state-of-the-art methods in the video captioning field (S2VT, SGC, and SCN) by a substantial margin. In particular, the best CIDEr score of V2CNet is $1.656$, while the CIDEr score of the closest runner-up SCN is only $1.561$. Furthermore, compared with the results by EDNet, V2CNet also shows a significant improvement in all experiments when different RNN types (i.e., LSTM or GRU) or visual features (i.e., VGG16, Inception, ResNet50) are used. These results demonstrate that by jointly learning both the fine-grained actions and the commands, the V2CNet can effectively encode both the visual features to generate the commands while is able to understand the fine-grained action from the demonstration videos. Therefore, the translation results are significantly improved.

Overall, we have observed a consistent improvement of our proposed V2CNet over EDNet in all variation setups. The improvement is most significant when the VGG16 feature is used. In particular, while the results of EDNet using VGG16 feature with both LSTM and GRU networks are relatively low, our V2CNet shows a clear improvement in these cases. For example, the Blue\_1 scores of EDNet using VGG16 feature are only $0.372$ and $0.350$ with the LSTM and GRU network respectively, while with the same setup, the V2CNet results are $0.391$ and $0.389$. We also notice that since EDNet only focuses on the translation process, it shows a substantial gap in the results between the VGG16 feature and two other features. However, this gap is gradually reduced in the V2CNet results. This demonstrates that the fine-grained action information in the video plays an important role in the task of translating videos to commands, and the overall performance can be further improved by learning the fine-grained actions.

From this experiment, we notice that there are three main factors that affect the results: the translation architecture, the input visual feature, and the external mechanism such as visual or action attention. While the encoder-decoder architecture is widely used to interpret videos to sentences, recent works focus on exploring the use of robust features and attention mechanism to improve the results. Since the IIT-V2C dataset contains mainly the fine-grained human demonstrations, the visual attention mechanism (such as in SGC architecture) does not perform well as in the normal video captioning task. On the other hand, the action attention mechanism and the input visual features strongly affect the final results. Our experiments show that the use of TCN network as the action attention mechanism clearly improves the translation results. In general, we note that the temporal information of the video plays an important role in this tasks. By extracting this information offline (e.g., from optical flow images as in S2VT, or with C3D network as in SCN), or learning it online as in our V2CNet, the translation results can be further improved.

\begin{figure*}
\centering
\footnotesize
  \stackunder[2pt]{\includegraphics[width=0.49\linewidth, height=0.09\linewidth]{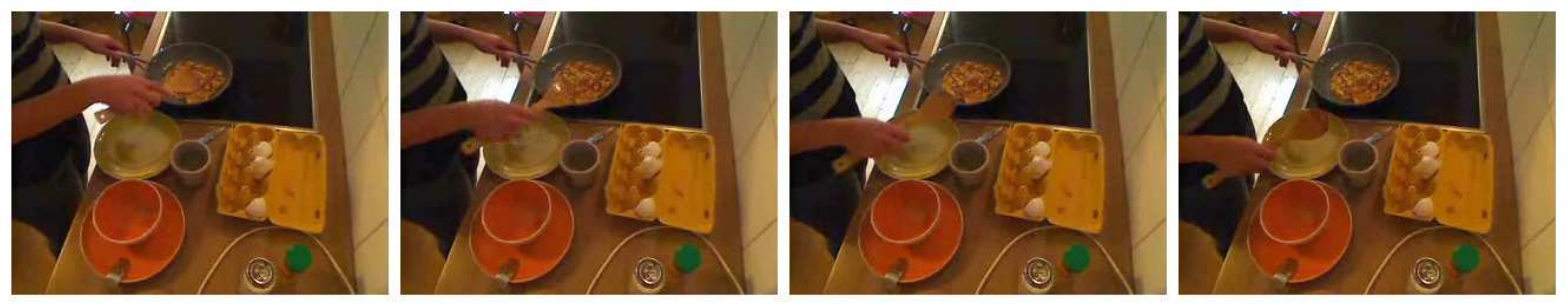}}  				
  				  {\tableCaption {righthand carry spatula} {righthand carry spatula} {lefthand reach stove} {lefthand reach pan} {righthand carry spatula} {righthand take spatula} }
  				  \vspace{2.5ex} 
  \hspace{0.25cm}%
  \stackunder[2pt]{\includegraphics[width=0.49\linewidth, height=0.09\linewidth]{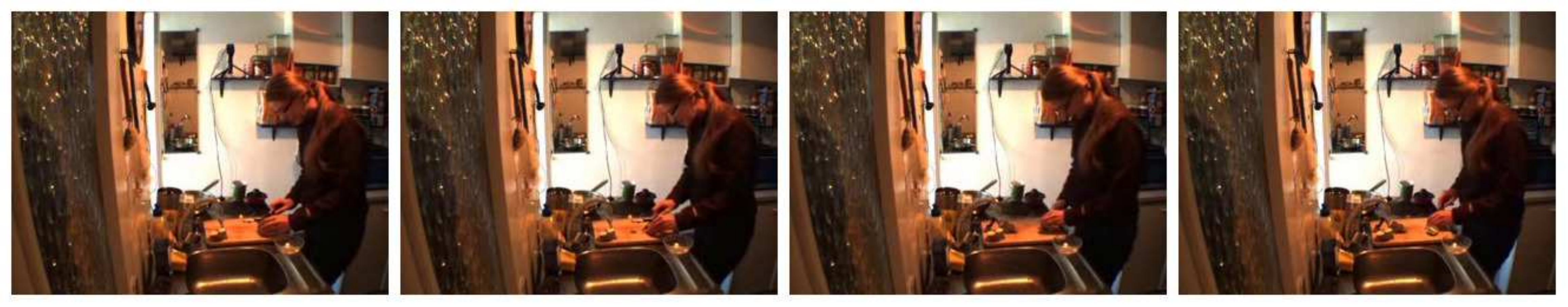}}  
  				  {\tableCaption {righthand cut fruit} {righthand cut fruit} {righthand cut fruit} {righthand cut fruit} {righthand cut fruit} {righthand cut fruit} }
  				  \vspace{2.5ex}
  \stackunder[2pt]{\includegraphics[width=0.49\linewidth, height=0.09\linewidth]{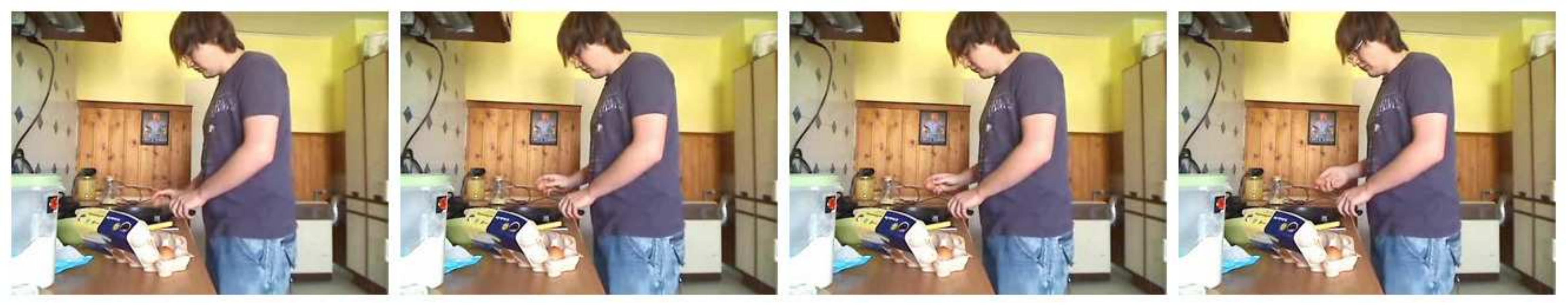}}  
    			  {\tableCaption {righthand crack egg} {righthand carry egg} {lefthand reach spatula} {righthand carry egg} {righthand crack egg} {lefthand carry egg} }
  				  \vspace{0ex}
  \hspace{0.25cm}%
  \stackunder[2pt]{\includegraphics[width=0.49\linewidth, height=0.09\linewidth]{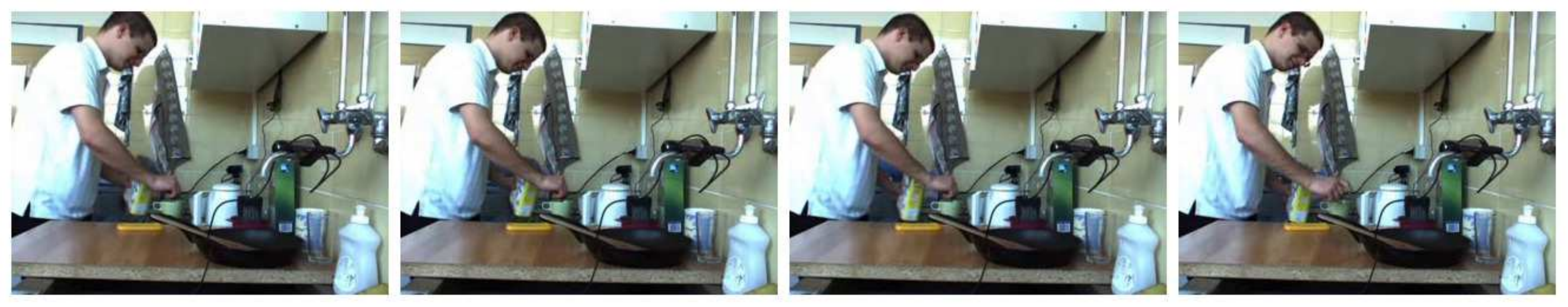}}    
				  {\tableCaption {righthand stir milk} {righthand hold teabag} {righthand place kettle} {righthand take cacao}{righthand stir coffee} {righthand stir milk} }
  				  \vspace{0ex}
  				  
  \stackunder[2pt]{\includegraphics[width=0.49\linewidth, height=0.09\linewidth]{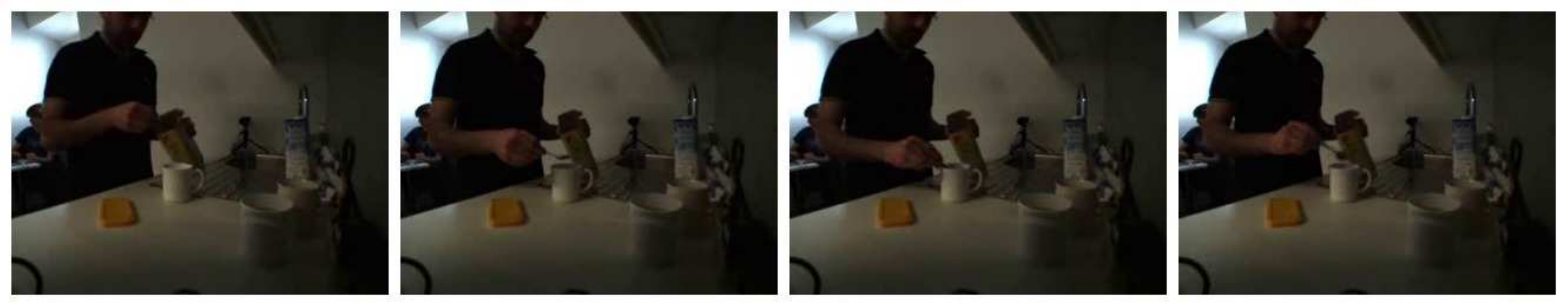}}  
    			  {\tableCaption {righthand transfer cacao to cup} {righthand reach cup} {righthand hold pan} {righthand pour cacao} {righthand transfer cacao to cup}{righthand transfer fruit} }
  				  \vspace{0ex}
  \hspace{0.25cm}%
  \stackunder[2pt]{\includegraphics[width=0.49\linewidth, height=0.09\linewidth]{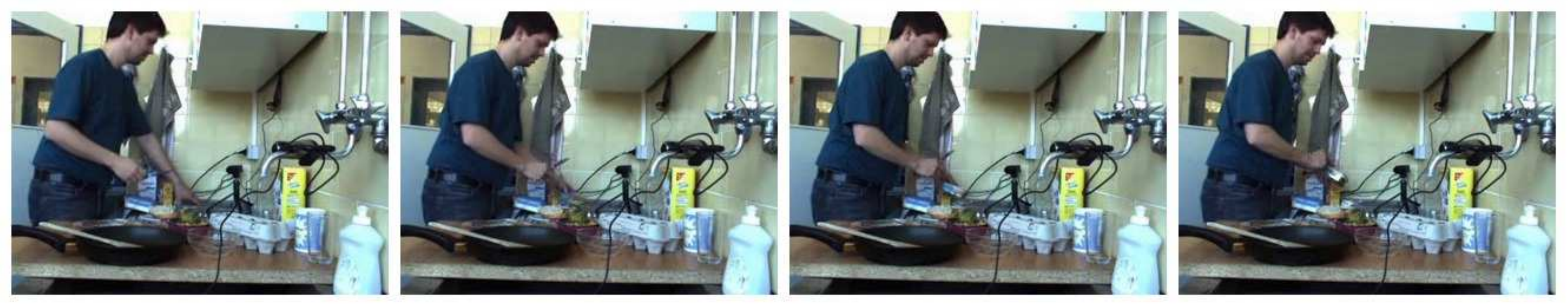}}    
				  {\tableCaption {lefthand take butter box} {lefthand reach teabag} {righthand place kettle} {righthand take kettle} {lefthand take topping}{lefthand reach milk bottle} }
  				  \vspace{0ex}
  
\vspace{1ex}
\caption{Example of translation results of our V2CNet and other methods in comparison with the groundtruth (\textbf{GT}).}

\label{Fig:main_result} 
\end{figure*}

Fig.~\ref{Fig:main_result} shows some comparisons between the commands generated by our V2CNet and other methods, compared with the groundtruth on the test videos of the IIT-V2C dataset. In general, our V2CNet is able to generate good predictions in many examples, while the results of other methods are more variable. In comparison with EDNet, V2CNet shows more generated commands with the correct fine-grained actions. We notice that in addition to the generated commands that are identical with the groundtruth, many other output commands are relevant. These qualitative results show that our V2CNet can further improve the translation results. However, there are still many wrong predictions in the results of all the methods. Since the IIT-V2C dataset contains the fine-grained actions, while the visual information is also difficult (e.g, the relevant objects are small and usually are covered by the hand, etc.). This makes the problem of translating videos to commands is more challenging than the normal video captioning task since the network has to rely on the minimal information to predict the command sentence.

To conclude, the extensive experiments using different feature extraction methods and RNNs show that our V2CNet successfully encodes the visual features and generates the associated command for each video. Our proposed method also outperforms recent state of the art by a substantial margin. The key idea that improves the translation results is the integration of the TCN network into the traditional encoder-decoder architecture to effectively learn the fine-grained actions in the video.

\subsection{Ablation Studies}

Although the traditional captioning metrics such as Bleu, METEOR, ROUGE\_L, and CIDEr give us the quantitative evaluation about the translation results, they use all the words in the generated commands for the evaluation, hence do not provide details about the accuracy of the fine-grained human actions. To analyze the prediction accuracy of the fine-grained actions, we conduct the following experiments: For both EDNet and V2CNet, we use the LSTM network and the visual features from the VGG16, Inception, and ResNet to generate both the commands from the translation branch, and the action class from the classification branch. The predicted actions of the translation branch are then automatically extracted by using the Stanford POS Tagger~\citep{Toutanova_tagger} to select the verb from the generated commands, while the classification branch gives us directly the fine-grained action class. For each variation of the networks, we report the success rate of the predicted output as the percentage of the correct predictions over all the testing clips. Intuitively, this experiment evaluates the accuracy of the fine-grained action prediction when we consider the translation branch also outputs the fine-grained action classes.

\newcolumntype{g}{>{\columncolor{Gray}}c}
\begin{table}[!ht]
\centering\ra{1.2}
\caption{The fine-grained action classification success rate.}
\renewcommand\tabcolsep{3.5pt}
\label{tb_result_action}
\hspace{2ex}

\begin{tabular}{@{}r|c|c@{}}  
\toprule 					 &
\small Feature 	& 
\small Success Rate  \\

\midrule
\multirow{3}{*}{EDNet~\citep{Nguyen_V2C_ICRA18}}  	&VGG16 		& 30.17\%     \\
												    &Inception	& 32.88\%     \\   
												    &ResNet50	& 32.71\%     \\									  
\cline{1-3}

\multirow{3}{*}{\makecell{V2CNet \\(translation branch)}} 	& VGG16 	& 31.75\%   \\
					   										& Inception & 34.41\%   \\
					   		    							& ResNet50 	& \textbf{34.81}\%   \\
\cline{1-3}					   			  
\multirow{3}{*}{\makecell{V2CNet \\(classification branch)}}& VGG16 	& 31.94\%   \\
					   										& Inception & 34.52\%   \\
					   		    							& ResNet50 	& 34.69\%   \\

\bottomrule
\end{tabular}
\end{table}

Table~\ref{tb_result_action} summaries the success rate of the fine-grained action classification results. Overall, we observe a consistent improvement of V2CNet over EDNet in all experiments using different visual features. In particular, when the actions are extracted from the generated commands of the translation branch in V2CNet, we achieve the highest success rate of $34.81\%$ with the ResNet50 features. This is a $2.1\%$ improvement over EDNet with ResNet50 features, and $4.64\%$ over EDNet with VGG16 features. While Table~\ref{tb_result_action} shows that the V2CNet clearly outperforms EDNet in all variation setups, the results of the translation branch and classification branch of V2CNet are a tie. These results demonstrate that the use of the TCN network in the classification branch is necessary for improving both the translation and classification accuracy. However, the overall classification success rate is relatively low and the problem of translating videos to commands still remains very challenging since it requires the understanding of the fine-grained actions. 

\subsection{Robotic Applications}

Similar to~\citep{Yang2015}, our long-term goal is to develop a framework that allows the robot to perform various manipulation tasks by just ``\textit{watching}" the input video. While the V2CNet is able to interpret a demonstration video to a command, the robot still needs more information such as scene understanding (e.g., object affordance, grasping frame, etc.), and trajectory planner to complete a manipulation task. In this work, we build a framework based on three basic components: action understanding, affordance detection, and trajectory generation. In practice, the proposed V2CNet is first used to let the robot understand the human demonstration from a video, then the AffordanceNet~\citep{AffordanceNet17} which is trained on IIT-AFF dataset~\citep{Nguyen2017_Aff} is used to localize the object affordances and the grasping frames. Finally, the motion planner is used to generate the trajectory for the robot in order to complete the manipulation tasks.

\begin{figure*}[ht]
  \centering
 \subfigure[Pick and place task using WALK-MAN]{\label{fig_resize_map_a}\includegraphics[width=0.99\linewidth, height=0.165\linewidth]{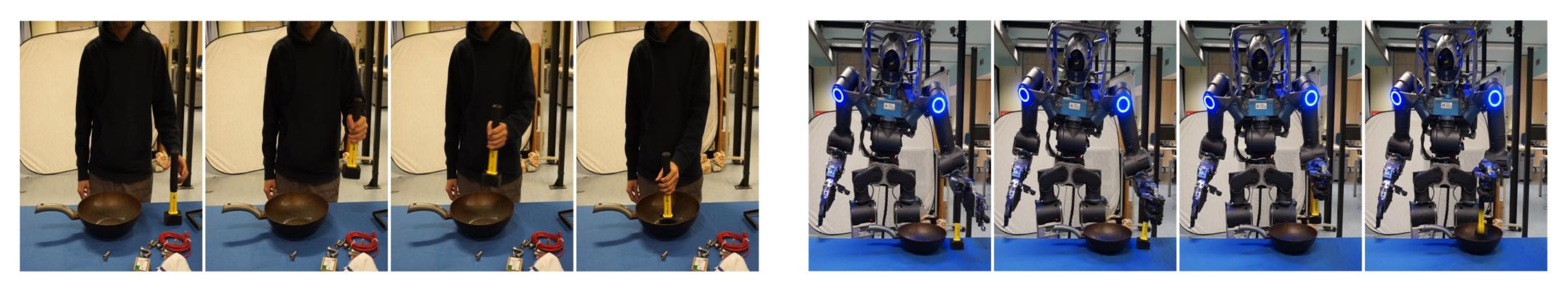}}
 \subfigure[Pouring task using WALK-MAN]{\label{fig_resize_map_b}\includegraphics[width=0.99\linewidth, height=0.165\linewidth]{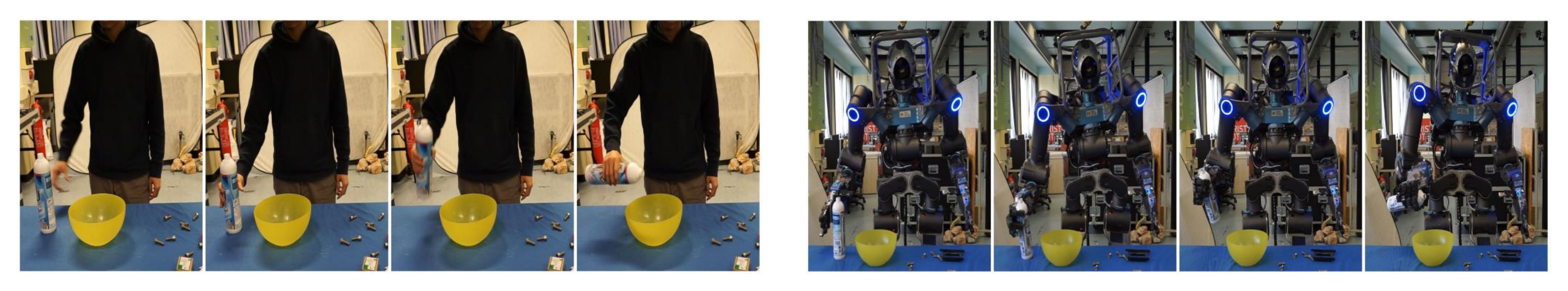}}
 \subfigure[Pick and place task using UR5 arm]{\label{fig_resize_map_b}\includegraphics[width=0.99\linewidth, height=0.165\linewidth]{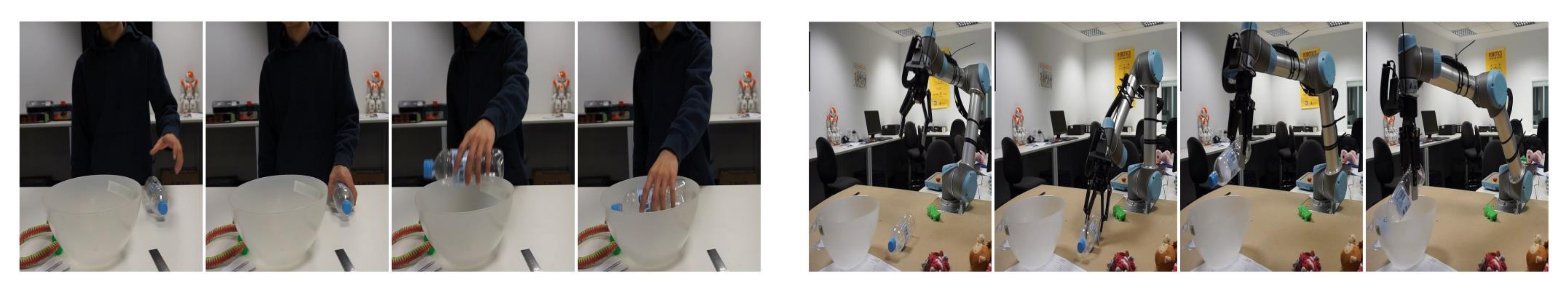}}

 \vspace{2.0ex}
 \caption{Example of manipulation tasks performed by WALK-MAN and UR5 arm using our proposed framework: \textbf{(a)} Pick and place task using WALK-MAN, \textbf{(b)} Pouring task using WALK-MAN, and \textbf{(c)} Pick and place task using UR5 arm. The frames from human instruction videos are on the left side, while the robot performs actions on the right side. We notice that there are two sub-tasks (i.e., two commands) in these tasks: grasping the object and manipulating it. More illustrations can be found in the supplemental video.}
 \label{fig_robot_imitation}
\end{figure*}

We conduct the experiments using two robotic platforms: WALK-MAN~\citep{Niko2016_full} humanoid robot and UR5 arm. WALK-MAN is a full-size humanoid robot with 31 DoF and two underactuated hands. Each hand has five fingers but only 1 DoF, and is controlled by one single motor. Therefore, it can only simultaneously generate the open-close grasping motion of all five fingers. We use XBotCore software architecture~\citep{muratore2017xbotcore} to handle the communication with the robot, while the full-body motion is planned by OpenSoT library~\citep{Rocchi15}. The vision system of WALK-MAN is equipped with a Multisense SL camera, while in UR5 arm we use a RealSense camera to provide visual data for the robot. Due to the hardware limitation of the end-effector in both robots, we only demonstrate the tasks that do not require precise skills of the fingers (e.g., cutting, hammering), and assume that the motion planning library can provide feasible solutions during the execution.

For each task presented by a demonstration video, V2CNet first generates a command sentence, then based on this command the robot uses its vision system to find relevant objects and plan the actions. Fig.~\ref{fig_robot_imitation} shows some tasks performed by WALK-MAN and UR5 arm using our framework. For a simple task such as ``righthand grasp bottle", the robots can effectively repeat the human action through the command. Since the output of our translation module is in grammar-free format, we can directly map each word in the command sentence to the real robot command. In this way, we avoid using other modules as in~\citep{Tellex2011} to parse the natural command into the one that uses in the real robot. The visual system also plays an important role in our framework since it provides the target frames for the planner. Using our approach, the robots can also complete long manipulation tasks by stacking a list of demonstration videos in order for the translation module. Note that, for the long manipulation tasks, we assume that the ending state of one task will be the starting state of the next task. Overall, the robots successfully perform various tasks such as grasping, pick and place, or pouring. Our experimental video is available at: \url{ https://sites.google.com/site/v2cnetwork}.

%% file: 5_discussion.tex
\section{Discussion} \label{Sec:discussion}
In this work, we divide the robot imitation task into two steps: the understanding step and the imitation step. We form the understanding step as a fine-grained video captioning task and solve it as a visual translation problem. From the extensive experiments on the IIT-V2C dataset, we have observed that the translation accuracy not only depends on the visual information of each frame but also depends on the temporal information across the video. By using the TCN network to encode the temporal information, we achieved a significant improvement over the state of the art. Despite this improvement, we acknowledge that this task remains very challenging since it requires the fine-grained understanding of human actions, which is still an unsolved problem in computer vision~\citep{lea2016learning}.

From a robotic point of view, by explicitly representing the human demonstration as a command then combining it with the vision and planning module, we do not have to handle the domain shift problem~\citep{yu2018one} in the real experiment. Another advantage of this approach is we can reuse the strong state-of-the-art results from the vision and planning fields. However, its main drawback is the reliance on the accuracy of each component, which may become the bottleneck in real-world applications. For example, our framework currently relies solely on the vision system to provide the grasping frames and other useful information for the robot. Although recent advances in deep learning allow us to have powerful recognition systems, these methods are still limited by the information presented in the training data, and cannot provide fully semantic understanding of the scene~\citep{AffordanceNet17}.

Due to the complexity of the whole robotic framework which requires many steps, our robotic experiments so far are qualitative. Since we design the framework as separated modules, the accuracy of the whole system can be improved by upgrading each module. From the robotic experiments, we notice that while the vision and planning modules can produce reasonable results, the translation module is still the weakest part of the framework since its results are more variable. Furthermore, we can also integrate the state-of-the-art LfD techniques to the planning module to allow the robots to perform more useful tasks, especially the ones that require precise skill such as ``cutting" or ``hammering". However, in order to successfully perform these tasks, the robots also need to be equipped with the sufficient end-effector. From a vision point of view, although our V2CNet can effectively translate videos to commands, the current network architecture considers each demonstration video equally and does not take into account the order of these videos. In real-world scenarios, the order of the actions also plays an important role to complete a task (e.g., ``putting on sock" should happen before ``wearing shoe"). Currently, we assume that the order of the sub-videos in a long manipulation task is known. Therefore, another interesting problem is to develop a network that can simultaneously segment a long video into short clips and generate a command sentence for each clip.

%% file: 6_conclusions.tex
\section{Conclusion}\label{Sec:con}
In this paper, we proposed V2CNet, a new deep learning framework to translate human demonstration videos to commands that can directly be used in robotic applications. Our V2CNet is composed of a classification branch to classify the fine-grained actions, and a translation branch to translate videos to commands. By jointly train both branches, the network can effectively encode both the spatial information in each frame and the temporal information across the time axis of the video. Furthermore, we also introduced  a new large-scale dataset for the video-to-command task. The extensive experiments with different recurrent neural networks and feature extraction methods showed that our proposed method outperformed current state-of-the-art approaches by a substantial margin, while its output can be combined with the vision and planning modules in order to let the robots perform different tasks. Finally, we discussed the advantages and limitations of our approach in order to address the work for the future.

\section*{Acknowledgment}
\addcontentsline{toc}{section}{Acknowledgment}
Anh Nguyen, Darwin G. Caldwell and Nikos G. Tsagarakis are supported by the European Unions Horizon 2020 research and innovation programme under grant agreement No 644839 (CENTAURO) and 644727 (CogIMon). Thanh-Toan Do and Ian Reid are supported by the Australian Research Council through the Australian Centre for Robotic Vision (CE140100016). Ian Reid is also supported by an ARC Laureate Fellowship (FL130100102). The authors would like to thank Luca Muratore for the help with the robotic experiments on WALK-MAN.